\documentclass[a4paper]{article}

\usepackage{INTERSPEECH2020}
\usepackage{multirow}

\title{Combining Prosodic, Voice Quality and Lexical Features\\ to Automatically Detect Alzheimer's Disease}
\name{Mireia Farr\'us, Joan Codina-Filb\`a}
\address{
  TALN Research Group, Department of Information and Communication Technologies \\
  Universitat Pompeu Fabra, Barcelona, Spain}
\email{mireia.farrus@upf.edu, joan.codina@upf.edu}

\begin{document}

\maketitle
\begin{abstract}
  Alzheimer's Disease (AD) is nowadays the most common form of dementia, and its automatic detection can help to identify symptoms at early stages, so that preventive actions can be carried out. Moreover, non-intrusive techniques based on spoken data are crucial for the development of AD automatic detection systems. In this light, this paper is presented as a contribution to the ADReSS Challenge, aiming at improving AD automatic detection from spontaneous speech. To this end, recordings from 108 participants, which are age-, gender-, and AD condition-balanced, have been used as training set to perform two different tasks: classification into AD/non-AD conditions, and regression over the Mini-Mental State Examination (MMSE) scores. Both tasks have been performed extracting 28 features from speech ---based on prosody and voice quality--- and 51 features from the transcriptions ---based on lexical and turn-taking information. Our results achieved up to $87.5 \%$ of classification accuracy using a Random Forest classifier, and $4.54$ of RMSE using a linear regression with stochastic gradient descent over the provided test set. This shows promising results in the automatic detection of Alzheimer's Disease through speech and lexical features.

\end{abstract}

\section{Introduction}

Dementia is a progressive syndrome characterized by a deterioration in cognitive function encompassing a severity spectrum, in which Alzheimer's Disease (AD) accounts about 70 \% of the cases. Its key symptoms, which can be manifested already in the early stages, are reflected by a loss of memory, a significant decline in thinking and reasoning skills, and language deterioration.

Language impairments related to AD are typically related to a progressive disruption in language integrity \cite{bayles1992relation}, with evidences of decline in performance at early stages \cite{mueller2018connected}. Language related features have been shown to capture the progression of language impairments through different clinical stages; therefore, they have been proposed by several authors as markers of disease progression \cite{ahmed2013connected,beltrami2018speech}. Some studies have also reported the ability of prosodic and other linguistic features to classify different cognitive stages from speech, showing promising approaches for the identification of preclinical stages of dementia \cite{konig2018use,wang2019towards}. Moreover, it has been shown that AD subjects produce fewer specific words, while syntactic complexity, lexical complexity or conventional neuropsychological tests are not altered, suggesting that subtle spontaneous speech deficits might occur already in the very early stages of cognitive decline \cite{verfaillie2019high, CompanyW18}.

Detection at early stages may help to foster prevention programs and delay the effects of dementia; therefore, an automatic detection of AD is crucial to slow down cognitive decline. In this light, spoken information is an accessible identifier that may help to develop non-intrusive and user-friendly applications. Although  several works have focused on the automatic detection and classification of AD \cite{fraser2016linguistic,luz2018method}, this field still suffers of a lack of standardisation. Moreover, many of the studies have traditionally focused on short speech samples under controlled clinical conditions rather than on spontaneous speech, which has been shown to be highly representative of speaker's characteristics \cite{farrus2008fusing}, becoming useful to detect individuals' impairments, and thus providing relevant information on AD symptoms \cite{luz2018method,luz2017longitudinal}.

The current paper is part of the ADReSS Challenge at Interspeech 2020, which defines a shared task based on spontaneous speech. Our work is based on the use of prosodic and voice quality features extracted from speech, as well as lexical and turn-taking information, tested over several machine learning classifiers. It includes a statistical feature analysis to see what speech and lexical features exhibit significant differences between AD and non-AD groups, as well as how they correlate with neuropsychological scores, and the two cognitive assessment tasks included in the ADReSS Challenge: the Alzheimer’s speech classification task and the neuropsychological score regression task.

The structure of this paper unfolds as follows. Section \ref{soa} briefly overviews the speech and lexical cues in Alzheimer's Disease; section \ref{setup} describes the experimental setup of our work; section \ref{results} presents the obtained results, and finally, discussion and conclusions are drawn in sections \ref{discussion} and \ref{conclusions}, respectively.

\section{Speech cues in Alzheimer's}\label{soa}

According to \cite{chien2019automatic}, systems relying on the speech signal to detect or assess Alzheimer's Disease can be classified into those using acoustic-dependent features, and those using context-dependent features. The former systems rely mainly on extracting speech-based features regardless of the linguistic content ---i.e., the transcriptions--- of the speech samples. These features can be related to the spectral characteristics, the voice quality, or the prosody of speech, among others. All these features can be used in a language-independent way. In turn, prosody is conveyed through the intonation, stress, and rhythm elements, which are most robust over noise and channel distortions than, for instance, spectral features \cite{atal1972automatic}. Moreover, rhythm features such as speech rate can also be estimated without the need of the corresponding transcriptions \cite{de2009praat}. On the contrary, systems relying on context-dependent features require the word transcriptions to infer lexical, semantic, syntactic, and grammar information, among others. When manual transcriptions are not available, they can be obtained through automatic speech recognition (ASR) systems. Linguistically-based features can be very informative to detect AD; however, they rely on the performance of an ASR system ---which are prone to introduce errors---, or human transcriptions ---which are very time-consuming---, and they are language dependant. The more elaborate the language analysis is, more sensible to ASR errors will be, making a real system less robust.

Several works have dealt with the use of acoustic information extracted from speech to detect AD status. \cite{luz2018method}, for instance, makes use solely of acoustic-dependent features from spontaneous spoken dialogues, including prosody-related rhythm features such as speech rate and turn-taking patterns; others focus their detection algorithms on spectral short-term features \cite{lopez2015automatic}, or jointly with prosodic and voice quality features to combine it with an emotional speech analysis \cite{lopez2013selection}. Voice quality features such as jitter and shimmer ---which account for cycle-to-cycle variations
of fundamental frequency and amplitude, respectively \cite{farrus2007jitter} --- and harmonicity, have also been used for Alzheimer's disease detection \cite{meilan2014speech,mirzaei2017automatic,mirzaei2018two}.

Apart from speech, other linguistic indicators based on the content transcriptions have been shown suitable for AD detection. Semantic, lexical and grammar deficits in general have been reported as useful indicators \cite{salmon1988lexical,altmann2001speech}. In \cite{fritsch2019automatic}, the authors compute the perplexity of transliterations in the spoken description of the same picture to distinguish between AD and non-AD conditions. Similarly, \cite{wankerl2017n} performs an AD automatic diagnose based on n-gram models with subsequent evaluation of the perplexity, also from the transcriptions of spontaneous speech. \cite{fraser2016linguistic} identifies several factors ---acoustic abnormality and impairments in semantics, syntax and information--- as linguistic markers in narrative speech to detect AD.

\section{Experimental setup}\label{setup}

\subsection{Data}

This work has been carried out over the dataset provided by the ADReSS Challenge \cite{bib:LuzHaiderEtAl20ADReSS}. The train and test sets consist of 108 and 48 participants, respectively, balanced in both gender and AD/non-AD conditions, and with age ranging from 50 to 80 years old. Age is also balanced in intervals of 5 years over male/female and AD/non-AD groups. Moreover, the train set provides the Mini-Mental State Examination (MMSE) scores, based on a 30-point scale, used to screen for dementia by estimating severity and progression of cognitive impairment \cite{folstein1975mini}.

All the participants were asked to record a spoken description of the Cookie Theft picture from the Boston Diagnostic Aphasia Exam \cite{piccini1995natural,
roth2011boston}. The average duration of all the files is 72 seconds. The corresponding transcriptions, annotated through the CHAT coding system \cite{macwhinney2000childes}, are also provided. A more detailed description of the ADReSS Challenge dataset can be found in \cite{bib:LuzHaiderEtAl20ADReSS}.

\subsection{Feature extraction}

\subsubsection{Acoustic features}

We extracted a set of acoustic features using the Praat speech software \cite{boersma2018praat}, which we classified as prosodic (relying on intonation, stress, and rhythm), and voice quality features. To account for intonation, we extracted the following log-scaled F0-based features (measured in Hertz), using the cross-correlation method with an interval of 3.3 ms and a Gaussian window of length 80 ms: \emph{mean F0} and its \emph{standard deviation}, \emph{max value of F0}, \emph{min F0}, \emph{range of F0}, \emph{F0 slope with octave jump removal}, and \emph{F0 slope without octave jump removal}. \emph{Mean intensity} (measured in dB) was also extracted to account for stress, and the following duration (in seconds) and rate features were extracted to account for rhythm: \emph{ratio of pauses}, \emph{avgerage pause length}, \emph{speech rate}, \emph{articulation rate}, \emph{average syllable duration}, and \emph{effective duration of the speech}. Duration and rate features were based on the algorithm proposed by \cite{de2009praat} without using the provided transcriptions to keep them as acoustic features measured in a language-independent way.

Voice quality features were also provided by Praat and included all jitter and shimmer measurements available (\emph{jitter\_loc}, \emph{jitter\_abs}, \emph{jitter\_rap}, \emph{jitter\_ppq5}, \emph{jitter\_ddp}, \emph{shimmer\_loc}, \emph{shimmer\_dB}, \emph{shimmer\_apq3}, \emph{shimmer\_apq5}, \emph{shimmer\_apq11}, and \emph{shimmer\_dda}), plus the following harmonicity-based features: \emph{harmonicity autocorrelation}, \emph{noise-to-harmonics ratio (NHR)}, and \emph{harmonics-to-noise ratio (HNR)}.

\subsubsection{Turn-taking and lexical features}

We extracted two types of features from the provided transcriptions: (1) the number of dialogue turns corresponding to the interviewer ---as an indicator of to what extent the subjects need further clarifications and stimuli---, min-max normalised, and (2) the fist occurrence of the 50 most relevant words. Since all subjects performed the description of the same picture, we can consider this task as a summarizing task. Word position in text has been been used in many systems as a relevance measure for summarization \cite{Hovy_97}. We expect that participants without any cognitive decline will catch the main ideas of the picture sooner, and their discourse will be structured from the most relevant to less relevant concepts. To produce a feature vector for all the users of a reasonable size, we only considered the 50 top most frequent nouns, adjectives and verbs found in the transcriptions of the training set. For these words we computed their score based on their first occurrence using the following formula:
\begin{equation}
    S(j,w_i)=1-\frac{min(turn(j,w_i),maxTurns)}{maxTurns}
\end{equation}
where $S(j,w_i)$ is the score of word $i$ for user $j$; $turn(j,w_i)$ indicates the first turn where word $i$ is found ($\infty$ if not found), and $maxTurns$ the maximum number of turns in the dataset.




\subsection{Analysis and prediction}

The first part of our study consists of a statistical analysis of our data. Given our sample size, we assume normality in the data distributions and we performed a one-way t-test with confidence level of $95 \%$ to analyse the statistical significance for each feature between control and AD groups, as well as the Pearson's correlation beetween MMSE and each of the features used (in the case of the lexical features, the average value over all 50 words was used).

The second part consists of the specific prediction experiments foreseen in the ADReSS Challenge. On the one hand, we perform an AD classification task, in which we build a model to predict the label for a speech session (AD or non-AD). On the other hand, we perform a MMSE score regression task, in which we predict each subject's MMSE score. To create our models, we used both speech and transcription data in both tasks, as well as gender and age information. The experiments with the training set were done with a 10-fold cross-validation setup. The results over the test set were provided by the Challenge organisers   \cite{bib:LuzHaiderEtAl20ADReSS} using the predictions we submitted.

Our classification task was performed through four different classifiers: Random Forest (RF), k-Nearest Neighbour (kNN), Support Vector Machines (SVM), and Multilayer Perceptron (MLP). For the regression task we used the following regressors: Random Forest (RF), Linear Regression with Stochastic Gradient Descent (LR-SGD), Support Vector Regression (SVR), and Multilayer Perceptron (MLP).

In both classification and regression tasks, RF and MLP were implemented with the Weka software \cite{witten2018practical} with its default parameters, except for the MLP, where learning rate was set up to $0.1$, and hidden layers to the wildcard value \emph{o} (i.e., a hidden layer with two neurons) in classification, and in regression when no lexical and turn-taking features were involved. For kNN, SVM, SVR and LR-SGD we used the Sklearn package in Python with the default parameters except: KNN (\emph{k=5}, weighted by distance) and SVM (SVC and SVR with polynomial kernel grade 3).  


\section{Results}\label{results}

\subsection{Statistical analysis}

Table~\ref{tab:statistics} shows the statistical analysis for both speech and lexical features between AD and non-AD groups. For the sake of simplicity, only those statistically significant features ($p < 0.05$) are shown.

\begin{table}[t]
  \caption{Statistical analysis of features between groups}
  \label{tab:statistics}
  \centering
  \begin{tabular}{lccc}
    \toprule
    \textbf{Feature} &  \textbf{non-AD} & \textbf{AD} & \textbf{p} \\
    \midrule
    mean F0 (log Hz) & $2.18$  & $2.23$ & $0.002$ \\
    std F0 (log Hz) & $0.13$  & $0.15$ & $0.049$ \\
    slope F0 (log Hz/s) & $29.18$  & $22.72$ & $0.003$ \\
    slope F0 w/o jump (log Hz/s) & $18.11$  & $15.03$ & $0.021$ \\
    jitter\_abs (ms) & $0.13$  & $0.10$ & $0.006$ \\
    pause duration (s) & $1.53$  & $1.99$ & $0.004$ \\
    speech rate (syll/s) & $2.07$  & $1.74$ & $0.015$ \\
    articulation rate (syll/s) & $3.80$  & $4.15$ & $0.019$ \\
    syllable duration (s) & $0.36$  & $0.25$ & $0.039$ \\
    lexical words & $0.28$ & $0.19$ & $< 0.001$ \\
    turn-taking interventions & $0.06$ & $0.16$ & $ < 0.001$ \\
    \bottomrule
  \end{tabular}
\end{table}

The Pearson's correlation revealed no significant relationship between MMSE and each of the prosodic and voice quality features. Only correlations $>0.2$ in absolute value were found with mean F0 (\emph{-0.29}), F0 standard deviation (\emph{-0.22}), and articulation rate (\emph{-0.25}). Instead, averaged lexical features showed up to \emph{0.59} of correlation with MMSE scores.

\subsection{AD classification task}

Table \ref{tab:classification1} shows the accuracy results obtained with several classifiers and combining different sets of features: prosodic (pros), voice quality (vq), lexical and turn-taking (lex), a selected (sel) set including those features that were shown to be statistically significant between non-AD and AD groups (Table \ref{tab:statistics}), and finally, a combination of all (all) features. Since prosodic features outperform the voice quality ones, the former were also combined with voice quality and lexical/turn-taking. Bold numbers show the five systems we submitted to the Challenge.





\begin{table}[ht]
  \caption{AD classification results (10-fold cross-validation).} 
  \label{tab:classification1}
  \centering
  \begin{tabular}{lcccc}
    \toprule
    \textbf{Features} &  \textbf{RF} & \textbf{kNN} & \textbf{SVM} & \textbf{MLP} \\
    \midrule
    pros & $0.657$ & $0.585$ & $0.640$ & $0.722$ \\
    vq & $0.602$ & $0.566$ & $0.629$ & $0.620$ \\
    pros + vq & $0.629$  & $0.595$ & $0.638$ & $0.713$ \\
    lex  & $0.694$  & $0.758$  & $0.797$ & $0.787$ \\
    lex + pros  & $0.667$ & $0.766$  & $\boldsymbol{0.825}$ & $0.796$ \\
    sel & $0.778$ & $0.748$ & $\boldsymbol{0.835}$ & $\boldsymbol{0.815}$ \\
    all & $\boldsymbol{0.787}$ &  $0.749$ & $\boldsymbol{0.844}$ & $0.787$ \\
    \bottomrule
  \end{tabular}
\end{table}

Table \ref{tab:classification2} shows the precision (P), recall (R), F1-score (F1) and accuracy (Acc) results obtained for the 5 submitted systems over the given test set. Results of the baseline provided in \cite{bib:LuzHaiderEtAl20ADReSS} are also shown for comparison. The best results obtained in each column are shown in bold.

\begin{table}[ht]
  \caption{Test results with the submitted classification systems.} 
  \label{tab:classification2}
  \centering
  \begin{tabular}{llcccc}
    \toprule
    & \textbf{class} & \textbf{P} &  \textbf{R} & \textbf{F1} & \textbf{Acc} \\
    \midrule
    \multirow{2}{*}{\emph{baseline}} & nAD & $0.67$ & $0.50$ & $0.57$ & \multirow{2}{*}{$0.625$}\\ 
    & AD & $0.60$ & $0.75$ & $0.67$ & \\
    \midrule
    \multirow{2}{*}{SVM all} & nAD & $0.85$ & $0.71$ & $0.77$ & \multirow{2}{*}{$0.792$} \\ 
    & AD & $0.75$ & \boldsymbol{$0.87$}  & $0.81$ & \\
    \midrule
    \multirow{2}{*}{SVM sel} & nAD & $0.86$ & $0.75$ & $0.80$ & \multirow{2}{*}{$0.812$} \\ 
    & AD & $0.78$ &  \boldsymbol{$0.87$} & $0.82$ & \\
    \midrule
    \multirow{2}{*}{SVM lex+pros} & nAD & $0.84$ & $0.67$ & $0.74$ & \multirow{2}{*}{$0.771$} \\ 
     & AD & $0.72$ & \boldsymbol{$0.87$} & $0.79$ & \\
    \midrule
    \multirow{2}{*}{MLP sel} & nAD & $0.86$ & $0.75$ & $0.80$ & \multirow{2}{*}{$0.812$} \\ 
    & AD & $0.78$ & \boldsymbol{$0.87$} & $0.82$ & \\
    \midrule
    \multirow{2}{*}{RF all} & nAD & $0.82$ & $0.96$ & $0.84$ & \multirow{2}{*}{\boldsymbol{$0.875$}} \\ 
    & AD & \boldsymbol{$0.95$} & $0.79$ & \boldsymbol{$0.86$} & \\
    \bottomrule
  \end{tabular}
\end{table}

\subsection{MMSE regression task}

\begin{figure}[ht]
  \centering
  \includegraphics[width=\linewidth, 
   trim=0 7 0 34, clip] {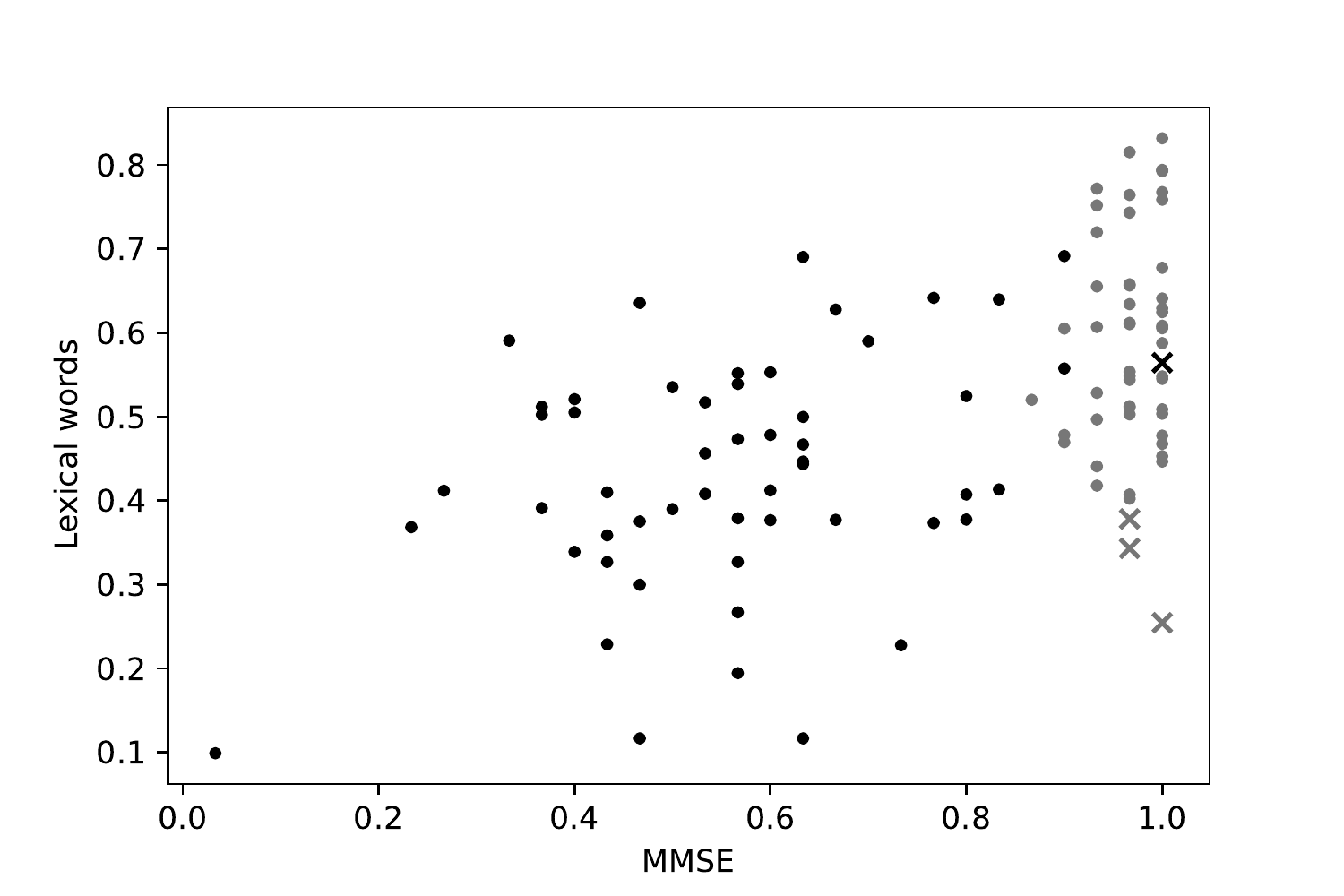}
  \caption{MMSE distribution against the 'lexical words' feature. Non-AD subjects in grey. Outliers marked as 'x'.}
  \label{fig:distribution}
\end{figure}

Table \ref{tab:regression1} shows the results obtained in terms of Root Mean Squared Error (RMSE) using the same feature combination as in Table \ref{tab:classification1}. Bold numbers show the systems submitted to the Challenge. For the outperforming system (LR-SGD with all features), a second system was trained after removing some subjects that were considered as outliers. Figure \ref{fig:distribution} shows the relationship between MMSE, AD/non-AD and the lexical richness of the users. Four of the subjects ---denoted by a cross--- were considered outliers: an AD subject with a MMSE score of 30, and three non-AD subjects that performed very poorly in the image description, reflected as low scores in the relevance of lexical words.

\begin{table}[ht]
  \caption{MMSE regression results (10-fold cross-validation).} 
  \label{tab:regression1}
  \centering
 \begin{tabular}{lcccc}
    \toprule
    \textbf{Features} & \textbf{RF} & \textbf{LR-SGD} & \textbf{SVR} & \textbf{MLP} \\
    \midrule
    pros   & $6.49$  & $ 8.03 $  & $ 6.84 $  & $7.60$ \\
    vq  & $7.32$   & $ 7.61 $  & $ 7.25 $  & $7.91$  \\
    pros + vq  & $6.57$  & $ 7.68 $  & $ 7.01 $  & $8.13$  \\
    lex   & $5.28$  & $ 5.00 $  & $ 5.33 $  & $7.14$  \\
    lex + pros  & $5.39 $  & $ 4.95 $  & $\boldsymbol{5.24}$  & $6.05$  \\
    sel  & $\boldsymbol{5.31}$  & $4.95$  & $5.26$  & $6.40$  \\
    all  & $5.55$  & $\boldsymbol{4.91/3.91}$  & $\boldsymbol{4.95} $  & $5.91$   \\
    \bottomrule
  \end{tabular}
\end{table}

Finally, Table \ref{tab:regression2} shows the RMSE values in the test set together with the baseline results provided by \cite{bib:LuzHaiderEtAl20ADReSS} for comparison.

\begin{table}[ht]
  \footnotesize
  \caption{Test results with the submitted regression systems.}
  \label{tab:regression2}
  \centering
  \begin{tabular}{cccccc}
    \toprule
     \multirow{2}{*}{\emph{\textbf{baseline}}} & \textbf{LR-SGD} & \textbf{SVR} & \textbf{RF} & \textbf{SVR} & \textbf{LR-SGD} \\
     & all & all & sel & lex+pros & - outliers \\
    \midrule
    $6.14$  & $4.65$  & $5.44$  & $4.83$ & $5.44$ & $\boldsymbol{4.54}$ \\
    \bottomrule
  \end{tabular}
\end{table}

\section{Discussion}\label{discussion}

A good understanding of the dataset characteristics and how different features behave under Alzheimer's Dementia condition is crucial to perform automatic prediction tasks. In this light, our statistical analysis reveals significant differences between non-AD and AD groups. In the case of prosodic features related to intonation, for instance, subjects with AD revealed higher values of F0 and its standard deviation, and lower values in F0 slopes. Although it seems there is no evident relationship between AD subjects and high F0 values, low slope values could signify a flatter intonation. In turn, lower jitter values could be an indirect effect from a lower F0 range. Looking at the rhythm element (durations and rates), AD subjects exhibit longer pauses in average and lower speech rates, though the articulation rate and syllable duration reveal that, although the AD subjects in general make longer pauses, the effective speech tends to be faster, maybe as an attempt to make up for the time they were silent. The position of relevant words is also significant, being them in more top positions within non-AD subjects and the only one significantly correlated with MMSE scores. The number of interventions from the interviewer is also significantly larger in the AD group.

Regarding the classification task, the best results were obtained with SVM, followed by MLP and RF, when using all the features or the selected set of features after a t-test calculation. Although Random Forest using all the features was our last option over the train set with 10-fold cross-validation, it appeared to be by far the outperforming system over the test set. However, it is also worth noting here the importance of having no AD false negatives ---i.e., non diagnosed subjects with AD---, given by the \emph{recall}, in which RF with all features is outperformed by the other four systems with a score of $0.87$.

Except for the kNN, the sole use of acoustic ---prosodic and voice quality--- features outperformed the baseline, especially in the case of the MLP classifier. Although in this case the performance only achieved up to $72.2\%$  of accuracy, such acoustic-based systems have the advantage of being language-independent, without relying on the need of human transcriptions or on the performance of an ASR system. Prosodic speaking rates and durations could have been obtained directly from transcriptions, getting a better estimation and probably higher accuracies, but at the expense of becoming a language-dependent feature.  In all cases, the use of lexical and turn-taking features outperformed those settings based only on acoustic features, although, in general, the combination of lexical and acoustic features gave better results.

The regression results are coherent with the classification ones in terms of features used: in the 10-fold cross-validation, prosodic features outperform the voice quality ones (except for LR-SGD classifier), and the combination of lexical and turn-taking with prosodic and voice quality features leads to better results. In all the systems submitted to the Challenge, we outperformed the baseline in terms of RMSE. The results presented also manifest the importance of a prior statistical analysis, since we found that after removing the MMSE outliers (see Figure \ref{fig:distribution}) the results of our ---so far--- best system improved.

\section{Conclusions and future work}\label{conclusions}

In this paper we have presented an automatic detection system for Alzheimer's Disease using several classification and regression algorithms and different sets of acoustic and linguistic features. With our best performing algorithms  we achieve up to $87.5\%$ of accuracy and $4.54$ of RMSE, by combining, in both cases, the whole set of prosody-related and voice quality features with lexical and turn-taking features. 

Acoustic-based systems have the advantage of being language-independent, without relying on the need of human transcriptions or on the performance of an ASR system. Although we extracted some of the features from the transcriptions, this could also be done without the need of human transcriptions or any ASR output: turn-taking could be obtained by means of speaker diarisation, while lexical words, once the system is trained, could be detected by word spotting.

Although previous studies had dealt with the use of acoustic- and context-dependent features, lexical features were usually based on perplexity of the transcriptions. To the best of our knowledge, this is the first work focusing on the relevant lexical words ---instead of perplexity---, and in combination with prosodic and acoustic features. The obtained results are promising and pave the way to improvements for further research. For instance, normalising the audio volume across all subjects' recordings to remove differences caused by the microphone placement ---as it was done in a segmented dataset also provided by the Challenge, but not suitable to compute long-term prosodic features---, removing external interviewer interventions prior to the acoustic-dependent feature extraction, a more accurate speech rate and duration calculation from transcripts ---and not relying only with acoustic estimations---, the inclusion of short-term spectral features for a more complete set of acoustic-dependent features, a more accurate analysis of feature normal distributions to select the best significance test for each feature, or the use of predicted MMSE values for the non-AD/AD classification task.

\section{Acknowledgements}

The first author has been funded by the Agencia Estatal de Investigación (AEI), Ministerio de Ciencia, Innovación y Universidades and the Fondo Social Europeo (FSE) under grant RYC-2015-17239 (AEI/FSE, UE).

\bibliographystyle{IEEEtran}

\bibliography{mybib}


\end{document}